\documentclass[sigconf]{acmart}

\usepackage[linesnumbered,ruled,lined]{algorithm2e}
\usepackage{algpseudocode}
\usepackage{multirow}
\usepackage{multicol}
\usepackage{array}
\usepackage{tabularx}
\usepackage{arydshln}
\usepackage{pifont}

\AtBeginDocument{%
  \providecommand\BibTeX{{%
    \normalfont B\kern-0.5em{\scshape i\kern-0.25em b}\kern-0.8em\TeX}}}

\setcopyright{acmcopyright}
\copyrightyear{2022}
\acmYear{2022}
\acmDOI{}

\acmConference[Asia CCS]{Proceedings of the 2022 ACM Asia Conference on Computer and Communications Security}{May 30th--June 3rd, 2022}{Nagasaki, Japan}
\acmBooktitle{Annual Computer Security Applications Conference, May 30th--June 3rd, 2022, Nagasaki, Japan}
\acmPrice{15.00}
\acmISBN{978-1-4503-XXXX-X/18/06}

\def\BibTeX{{\rm B\kern-.05em{\sc i\kern-.025em b}\kern-.08em
    T\kern-.1667em\lower.7ex\hbox{E}\kern-.125emX}}

\begin{document}

\title{ABC-FL: Anomalous and Benign client Classification in Federated Learning}

\author{Hyejun Jeong}
\email{june.jeong@skku.edu}
\affiliation{
  \institution{Sungkyunkwan University}
  \city{Suwon}
  \country{Republic of Korea}
}


\author{Joonyong Hwang}
\email{brian0316@skku.edu}
\affiliation{%
  \institution{Sungkyunkwan University}
  \city{Suwon}
  \country{Republic of Korea}
}

\author{Tai Myung Chung}
\email{tmchung@skku.edu}
\affiliation{%
  \institution{Sungkyunkwan University}
  \city{Suwon}
  \country{Republic of Korea}}

\renewcommand{\shortauthors}{Jeong et al.}

\begin{abstract}
  Federated Learning is a distributed machine learning framework designed for data privacy preservation i.e., local data remain private throughout the entire training and testing procedure. Federated Learning is gaining popularity because it allows one to use machine learning techniques while preserving privacy. However, it inherits the vulnerabilities and susceptibilities raised in deep learning techniques. For instance, Federated Learning is particularly vulnerable to data poisoning attacks that may deteriorate its performance and integrity due to its distributed nature and inaccessibility to the raw data. In addition, it is extremely difficult to correctly identify malicious clients due to the non-Independently and/or Identically Distributed (non-IID) data. The real-world data can be complex and diverse, making them hardly distinguishable from the malicious data without direct access to the raw data. Prior research has focused on detecting malicious clients while treating only the clients having IID data as benign. In this study, we propose a method that detects and classifies anomalous clients from benign clients when benign ones have non-IID data. Our proposed method leverages feature dimension reduction, dynamic clustering, and cosine similarity-based clipping. The experimental results validate that our proposed method not only classifies the malicious clients but also alleviates their negative influences from the entire procedure. Our findings may be used in future studies to effectively eliminate anomalous clients when building a model with diverse data.
\end{abstract}




\maketitle

\section{Introduction}
\label{section1}

Federated Learning (FL) \cite{mcmahan2017communication} was introduced in 2017 to deal with the rising issues related to data privacy. Despite great advancements in Deep Learning (DL), the existing DL techniques are difficult to implement in a privacy-preserving way. Canonical Machine Learning (ML) or DL methods require that data be placed in a single location to train the model, giving rise to two possible issues; information disclosure and the creation of a single point of failure. On the other hand, FL, which uses DL in a collaborative and distributed manner, prevents the raw data from being transmitted and collected into a single central server, making it possible to exploit the strengths of DL while preserving data privacy.

Therefore, FL provides data privacy by its design in contrast to the traditional centralized DL approach and the classical distributed learning approach. The conventional DL methods collect all the needed data into a central server or a physical data-center \cite{roh2019survey}; and distributed learning \cite{balcan2012distributed} distributes an Independently and Identically Distributed (IID) raw dataset into multiple local devices to train in a decentralized way. The FL setting restricts each local client and global server from accessing the raw data of other clients. Private training data permanently resides in the local machine instead of being collected to a centralized location. Only the model updates, such as the weights, are communicated. Furthermore, the global server does not need to store the participating individual clients' model updates, to update a global model, as the updates are for one-time-use. This characteristic of FL resolves the pressing concerns about data privacy and security, thereby reducing the attack surface \cite{mcmahan2017communication}.

A vanilla FedAvg \cite{mcmahan2017communication} works by repeating the following steps. A server initializes a global model and advertises it to all the participating clients. Each client trains the model with their privately-owned data and then uploads the model updates to the global server. The global server aggregates the model updates by averaging them. These steps are repeated until a stopping criterion is met. The fact that the clients only keep their training data locally instead of transmitting them to a single location functions as a beneficial factor not only for the preservation of data privacy but also the efficiency of the entire framework.

As FL inherits the vulnerabilities of DL, it is especially vulnerable to data or model poisoning attacks \cite{bonawitz2019towards, fang2020local, bagdasaryan2020backdoor, cao2020fltrust}. Data poisoning attacks manipulate the training data aiming to harm the performance of the model \cite{shen2016auror}, while model poisoning attacks corrupt the model parameters. Model poisoning attacks are further divided into untargeted and targeted attacks. The untargeted attacks tamper with the global model so as to have it generally make wrong predictions, while targeted attacks target attacker-chosen inputs, resulting in the model making attacker-chosen predictions, and untargeted inputs remain intact. Backdoor attacks are a type of targeted attack that stealthily trains the model on backdoor task(s) without influencing the main tasks \cite{chen2017targeted, gu2019badnets, liao2018backdoor}. In this study, we focused on a targeted model poisoning attacks.

FL is also susceptible to non-IID participation, similarly to DL. In a real-world scenario, it is hard to expect all data to be distributed independently and identically. Instead, they are generally heterogeneously distributed; for example, different number of data samples for each class, distinct classes of data for each client, different features but with the same label, or the same feature but with different labels \cite{kairouz2019advances}. Since most ML or DL frameworks are built on top of the assumption that the data are in IID, the performance typically degrades when we apply such models to non-IID data \cite{kairouz2019advances, zhao2018federated, sattler2019robust}. 

Compared to normal ML or DL techniques where the centralized server can investigate the raw data, neither the server nor the other clients are able to access the local data in FL. The inaccessibility to the private data, comes the difficulty of differentiating the clients having non-IID data from those than contain malicious data. Unfortunately, in practical applications, the dataset usually tends to be non-IID, which complicates efforts in anomaly detection than that of a single IID set \cite{herlands2018gaussian}. Non-IID data is extraordinarily complicated, having non-trivial relationships across time, space, and other variables. Due to the non-independent nature of non-IID data, the level of deviation of each data point, from IID, diminishes the ability to discover subtle anomalies. If these data points are treated as IID, crucial co-variance structures are ignored, resulting in overestimated detection of anomalousness. Nevertheless, identifying and filtering out the maliciousness of the participating clients is essential due to the vulnerable nature of the FL framework.

Various previous studies on anomaly detection in FL do not involve non-IID data, which is the most practical scenario, but only IID and malicious data \cite{li2020learning, qin2020selective, tolpegin2020data, bagdasaryan2020backdoor, cao2020fltrust, fang2020local}. Other works \cite{briggs2020federated, sattler2019robust} aiming for performance improvement over non-IID data incorrectly define label-flipped (malicious) data as non-IID data. However, few studies on identifying and classifying homogeneous, heterogeneous, and malicious data were done under the FL environment. 

Thus, we propose an anomalous and benign client classification method in FL called \textit{ABC-FL}. The proposed method leverages adaptive clustering based on 2-dimension Principal Component Analysis (PCA) and the cosine similarity between each local and global model's weight vectors. The results of this study show that the proposed method effectively classifies the malicious participating clients when the clients equipped with non-IID data exist along with the IID data. Also, compared to the baseline, the proposed method notably decreases the backdoor success rate. \\

Our main contributions can be summarized as follows:
\begin{itemize}
    \item We propose an anomalous client detection method where the server is neither pre-trained nor knowledgeable about the clients. 
    \item We propose a method that decides whether a client is malicious based on the unit of a cluster but not on the individual client. 
    \item We propose an Anomalous and Benign client Classification model in FL setting (ABC-FL), that considers IID, non-IID, and malicious clients. 
\end{itemize}

\section{Related Works \& Background} 
\label{section2}

\begin{figure*}
    \centering
    \includegraphics[width=\linewidth]{"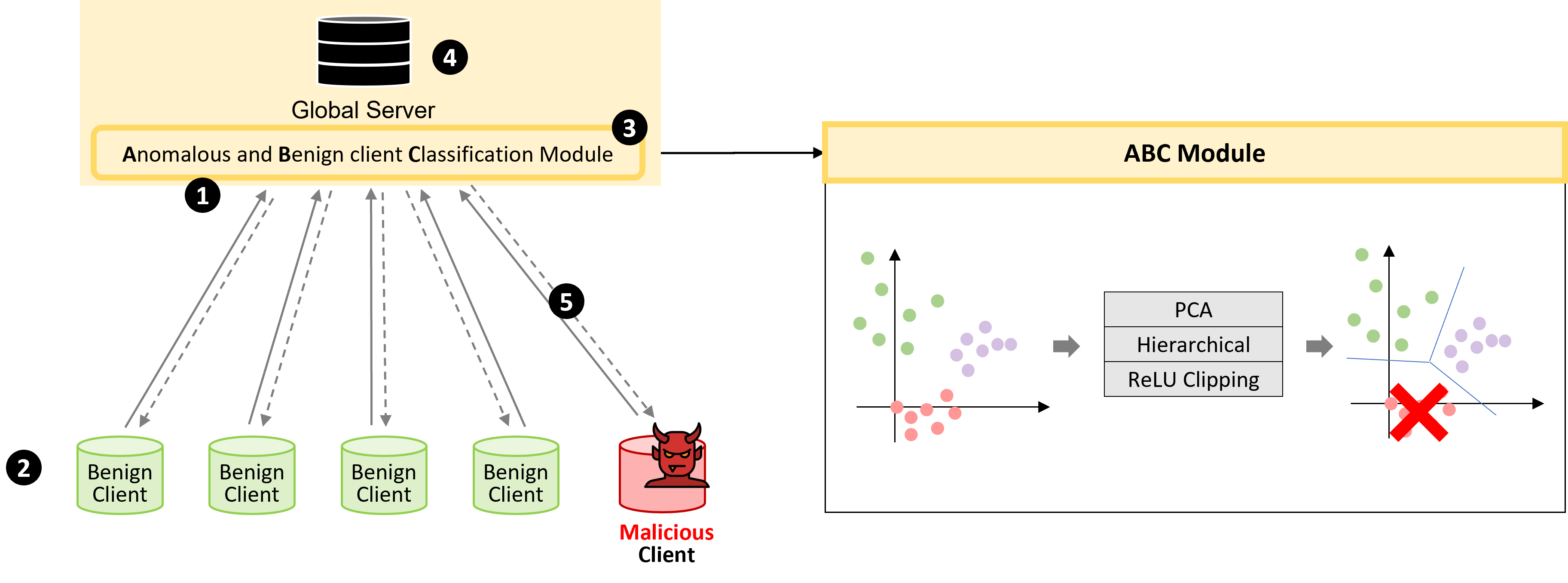"}
    \caption{The Overall Mechanism of our ABC Module inserted FL (ABC-FL). 1) The global server sends initialized global model to all clients. 2) Each local model is trained with its private data. 3) An ABC module identifies the anomalous clients. 4) The clients' weights are aggregated except for the malicious ones. 5) The server sends the updated model to all clients. Note that each benign client has either an IID or non-IID data distribution.}
    \label{fig:overview}
\end{figure*}

\subsection{Data Heterogeneity in Federated Learning}

One important principle of FL is that the global server cannot directly verify each participating client's data distribution. Various studies have been conducted with regards to the data heterogeneity issue in FL. ~\cite{zhao2018federated} introduced the concept of sharing a small portion of data that the global server owns. However, its shortcoming is that it breaches the fundamental concept of FL that no raw data must be shared at any point in time. FedAdam, FedYogi ~\cite{zaheer2018adaptive}, and FedAdagrad ~\cite{reddi2020adaptive} applied adaptive optimization algorithms for the server-side optimizer. ~\cite{sattler2019robust, briggs2020federated} leveraged a hierarchical clustering algorithm. They built specialized models for each cluster of clients based on the cosine similarity between their weight updates. However, none of the works mentioned simultaneously consider the malicious clients but only IID and non-IID data.

Non-IID can be decoupled into two distributions: non-identical data distribution and non-independent (dependent) data distribution. Under the non-IID setting, the data distribution of participating clients are different. 
The non-identically distributed data can be broadly divided into five categories, by how the data distributed: feature distribution skew, label distribution skew, concept drift, concept shift, and quantity skew \cite{kairouz2019advances, moreno2012unifying}. We give a brief explanation on the five categories of non-IID below:
\begin{itemize}
    \item Feature distribution skew (Covariate Shift): Marginal feature distributions vary across clients even if the labels of given distinct features are the same. For example, multiple individuals’ drawings of the same object might have different styles. 
    \item Label distribution skew (Prior Probability Shift): Marginal label distributions vary across clients even if the features of given labels are the same. For example, some clients have limited access to data (e.g., tied to particular geolocations) and the label distribution may differ across clients. 
    \item Concept drift (Same Label, Different Features): Conditional feature distributions given labels vary across clients, even if the marginal label distributions are the same. For example, two photos of a car taken in summer and winter at the same place. The feature distribution given labels could be different (imagine a snow-covered car in the winter) while both photos are of the same car.
    \item Concept shift (Different Label, Same Features): Conditional label distributions given features vary across clients, even if the marginal feature distributions are the same. Google’s Gboard is an exact example of this type of non-identicality. The next predicted word varies by each user while the preceding word is the same.
    \item Quantity skew: Amounts of data vary across clients.
\end{itemize}

Additionally, most of the aforementioned works assume a heterogeneous data environment. For instance, each client has only a subset of classes of the total data, such as having only two classes of data out of ten classes of the entire data (label distribution skew). In this study, we prepare our non-IID dataset by having different domains of the same data class (concept drift). Details of our dataset preparation are discussed in Section 3.1.

\subsection{Defenses against Anomalous Attacks in Federated Learning}
Previous works tried to build a robust model by excluding malicious clients before aggregating participants' parameters to protect the global model against backdoor attacks in FL. There are various approaches to identify malicious clients. Here, we investigate the previous works in three categories. \\

\noindent\textbf{Defenses using PCA. }
 \cite{tolpegin2020data, pillutla2019robust, chatterjee2021federated} proposed a Principal Components Analysis (PCA) based method to detect malicious clients for different levels of data corruption. They collected and projected each client's weight updates to a lower dimension. They showed that the weight distributions of benign and malicious clients are clearly distinguishable with fewer features. Specifically, \cite{chatterjee2021federated} claimed that the projection to the 2-D shows the negligible difference to that of a higher dimension; and that there is no reason to choose a higher dimension that requires more computation and memory. However, the previous works are limited in that the local datasets are either in IID or adversarial fashion, which is commonly not the case in the real world. They also disregard the base setting that non-IID samples exist along with the malicious samples. \\

\noindent\textbf{Defenses using Thresholding. }
 \cite{sun2019can} proposed norm-based thresholding. They calculate the L2 norm value of local weights and compare it with a specified threshold. Local clients are omitted from the aggregating step if the amount is below the threshold. \cite{cao2020fltrust} proposed FLTrust, which weighs the local clients' parameters based on a trust score. The trust score is a ReLU clipped cosine similarity, retrieved by applying ReLU clipping on the cosine similarities between each local model and the global model's parameters. The ReLU clipping step converts the trust score to zero for clients with negative values, and ultimately excludes the clients with negative similarity score. Their method assumed that the server has a small portion of the innocent dataset which may violate the zero-knowledge assumption of the server in FL. Similar to the above limitation, these works did not consider the non-IID setting. \\

\noindent\textbf{Defenses using Clustering. }
 \cite{nguyen2021flguard} proposed FLGuard, that deploys a dynamic clustering algorithm to identify malicious clients. They used HDBSCAN on the model updates based on the pairwise cosine distance between all clients. Their work, however, is limited in that they considered only IID settings. \cite{yu2020towards} proposed a group-wise aggregation approach to address data heterogeneity, but not to defend against attacks. They developed a clustering algorithm on model parameters to group them so that if a new client comes in, its cluster assignment is determined by estimating the average center of the cluster. However, this work only dealt with the non-IID issue, but not defense against suspicious activities. \\
 
Our work is different from the aforementioned works in that we propose a method that leverages a dynamic clustering approach to mitigate adversarial attacks. In our work, dynamic clustering with cosine similarity is leveraged to identify malicious clients while the server has zero knowledge about the clients, thereby strictly adhering to the fundamental premise of FL.

Taking into consideration the aforementioned limitations in previous works, we note that a precise classification even with the participants having non-IID data distributions is one of the essential considerations that should be made to ensure data diversity and system robustness. In this work, we consider both samples with IID and non-IID distributions as benign against malicious samples. Hence, we consider three types of clients: an anomalous client with malicious samples and two types of benign clients with IID and non-IID samples.

\section{Anomalous and Benign client Classification in FL}

\label{section3}

\begin{algorithm}
\DontPrintSemicolon
{
Initialize $w_0$ \tcp*{Run on the Server}
  \For {each round $t = 1, 2,...$}{
    \For{each client $k \in K$}{
        $w_{t+1}^k \leftarrow$ \tt{ClientUpdate}($k, w_t$)\;
    }
    \For{every three rounds}{
        $D \leftarrow$ \tt{ABCModule}($K$, $w_{t+1}^k$, $w_{t}$)\;
    }
    \For{each client $(k \in K)$ \tt{and not in} $D$ $(k \not\in D)$}{
        $w_{t+1} \leftarrow \sum_{k=1}^{K} \frac{n^k}{n}w_{t+1}^k$ \;
    }
} 
\;

\SetKwFunction{FClientUpdate}{ClientUpdate} 
\SetKwFunction{FABC}{ABCModule}

\SetKwProg{Fn}{Function}{:}{} 
\Fn{\FClientUpdate{$k$, $w$}}{
    $ \beta \leftarrow$ \textit{split $P_k$ into batches of size $B$} \\
    \For {each local epoch $i = 1, 2, ...,E$}{
        \For{batch $b \in \beta$}{ 
        $w \leftarrow w - \eta\Delta_i(w;b)$ \;
        }
    }
    \KwRet $w$\;
  }
\;

\Fn{\FABC{$K$, $W_{local}$, $W_{glob}$}}{
    \tt{PCA}($W_{local}$)\; 
    $C \leftarrow$ \tt{HierarchicalClustering}($W_{local}$) \;
    $S \leftarrow$ \tt{CosineSimilarity}($W_{local}$, $W_{glob}$) \;
    \For{each cluster $c \in C$}{
        \For{each client $k \in c$}{
            \If{$S_c <= 0$ or $S_c <= \sum\frac{S}{n}$}{$count \leftarrow count + 1$\;}
        }
        \If{\tt{length($c$)/}$2$ <= \tt{count}}{
            \tt{detected} $\leftarrow$ \tt{detected + }$c_k$\;
        }
    }
    \KwRet detected } 
}
\parbox{\linewidth}{\caption{ABC-FL. The $K$ clients are indexed by $k$, and $n$ is the total number of clients; $B$ is the local batch size, $E$ is the number of local epochs, and $\eta$ is the learning rate. $W_{local}$ and $W_{glob}$ are a list of client models' weights and the global model's weights, respectively. $C$ is a set of clusters, indexed by $c$, and $S$ is the list of similarities between each local client's weights $w_{l}^k$. Note that \textit{ClientUpdate} is run on each client $k$, and \textit{ABCModule} is run on the server.}}
\label{alg:algorithm}
\end{algorithm}

Anomalous and Benign client Classification module in FL (ABC-FL) is a defense against backdoor attacks. Some clients have IID data, others have non-IID data, and others have malicious data. Our ABC-FL complies with a fundamental premise in the FL: no access to raw data. ABC-FL also prohibits the server from having any knowledge of the clients' data. As such, the detection is purely reliant on the weight vectors of the global model and the clients' local models. 

ABC-FL intervenes in the FL training process to identify malicious clients; the ABC module inspects the weight changes of each local model before the weight updates are aggregated on the server. Then the identified malicious client's weight updates are dropped so that the server does not use the malicious clients' weights to update the global mode as illustrated in Figure \ref{fig:overview}. 

The ABC Module consists of three stages: 1) Principal Component Analysis (PCA), 2) Hierarchical Clustering, and 3) Cosine Similarity-based ReLU Clipping. The module begins by applying PCA on the clients models' weights to reduce the number of dimensions into two. It then leverages hierarchical (agglomerative) clustering on the reduced model parameters without requiring the user to specify the number of clusters in advance. The cosine similarity between each local model's and global model's parameters are then computed. The ABC module decides that a client is anomalous if the cosine similarity is negative or smaller than the mean of all clients' similarities. If more than half of the clients in a cluster are identified as anomalous, the clients in the cluster are all treated as anomalous. This leads to the weights in the entire cluster to be excluded from aggregation. \\

\noindent
\textbf{PCA.}
PCA is used to preprocess participating clients' local models' weights. PCA is typically used for the prevention of overfitting, therefore, the technique is used to relax the trivial non-IID pattern in the clients' weight updates. In our work, we utilized PCA to analyze the weights of each client at a specific epoch \cite{tolpegin2020data}. Suppose we consider a case where it takes 50 rounds to train an FL system. During each pre-defined epoch (e.g., every other epoch), the weights of participating clients are collected and used to compute a difference between them and the weights of the global server during the preceding epoch.
After acquiring the differences between the participating clients and the global server, we employ PCA to decrease the dimension to a 2-D representation. PCA, however, is only applied to the weights after the fully connected layer to relax the computational complexity. \\

\noindent
\textbf{Hierarchical Clustering.}
Agglomerative clustering is used to group the clients based on the PCA-reduced model's parameters. As the server does not know the client composition (i.e., IID, Non-IID, and/or malicious), dynamic clustering would be more effective in properly gathering the clients. Instead of specifying the number of clusters a priori, hierarchical clustering adaptively determines the number of clusters. This means that the clients can be clustered into more than three clusters, which will be examined by an individual cluster of clients. We set a distance threshold instead of defining the number of clusters. The threshold is set by calculating the inner dot product between each local and global models' weights, computing the mean and standard deviation of the dot products, and then subtracting the mean by the standard deviation. We set the linkage as complete, in which clusters are grouped based on the maximum value between the two clusters, and an L2 norm as a distance metric so as to cover both magnitude and the direction at different stages. Note that the linkage and distance metric can vary, depending on the dataset. The clusters of clients are further evaluated by the cosine similarity. \\

\noindent
\textbf{Cosine Similarity and ReLU Clipping.}
The final and most significant stage is classifying malicious clients based on their cosine similarity with legitimate clients. According to \cite{cao2020fltrust}, the cosine similarities between the weights from the global model and each participating local client are computed and compared for each cluster of clients. Specifically, we count the number of clients in which the similarities are negative or less than the mean of the total set of similarities. If the counted number is greater than half of the size of the cluster (i.e., the number of clients in the cluster), the cluster is designated as malicious and all clients belonging to that cluster are categorized as malicious. The reason beyond this is as the precision of clustering often exceeds that of the malicious detection. For instance, some data in the malicious-major cluster may be falsely detected as benign, yet this algorithm can filter such mis-prediction. All the clusters of clients classified as malicious are concatenated to prevent the server from averaging their model parameters. As a result, the global model is updated without the influence of malicious clients.

\subsection{Implementation Detail }
We adopted FedAvg \cite{mcmahan2017communication} to aggregate the weights of benign clients. FedAvg works by updating each client once on the current model using its private data, and then the server averages all clients' updates. The proposed ABC module sits on the global server, allowing for the module to prevent malicious local model parameters from being aggregated and used to update the global model. We term this approach \textit{ABC-FL}, and the algorithm is as shown in Algorithm 1.

The global server initializes a global model and sends its copy to all local clients. Each client updates their local models only with the locally owned data and then sends their updated weights to the server. The ABC module then identifies the anomalous clients for every three rounds to avoid initial misclassification from being solidified. For each time our ABC module engages, only the updates of benign clients are averaged, while the malicious clients' model parameters are ignored. We exclude malicious clients to prevent their weights from influencing the global model but choose not to drop them for further training owing to the possibility of them being false positives. Ideally, the server would reject presumably normal but indeed malicious clients while aggregating the seemingly evil but innocuous clients.

We also assume that the model should be light enough to run on the local devices, which typically have little computing power. For this reason, we choose to use a simple two-layer neural network composed of two convolutions and two fully connected layers.



\section{Experiment}
\label{section4}

\begin{figure*}
    \centering
    \includegraphics[width=0.8\linewidth]{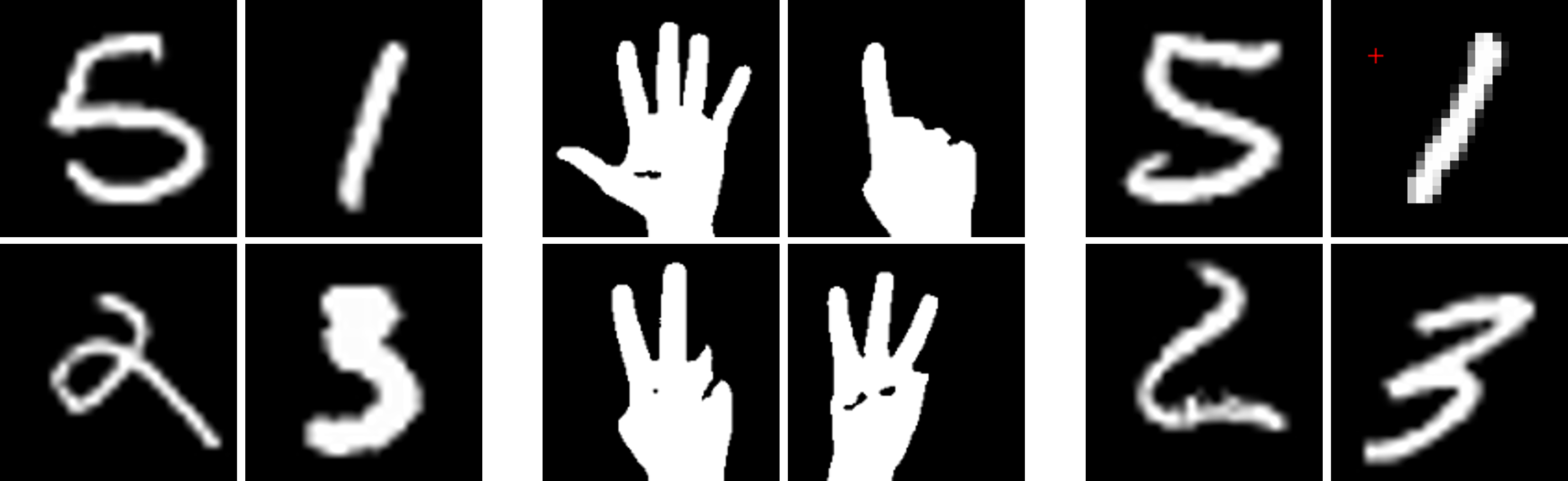}
    \caption{An example of sample dataset used for our experiments. IID, non-IID, and malicious sample images from left to right. The IID clients only have IID data samples, while non-IID clients have different proportion of non-IID and IID samples. The malicious clients also have a mixture of the different proportion of malicious and IID samples.}
    \label{fig:sample-data}
\end{figure*}

Ahead of experiments, we made two assumptions:
\begin{itemize}
    \item Two types of benign clients having IID and non-IID data, respectively, participate in the FL training procedure.
    \item Less than half of the participating clients are malicious clients having backdoor samples. 
\end{itemize}

To address all possible scenarios, We gradually increased the fraction of participating malicious clients from 0.1 to 0.4 with a step of 0.1 and the degree of non-IID from 0.2 to 1.0 with a step of 0.2. We evaluated the efficiency of our proposed method in IID and malicious (IID+MAL) and non-IID and malicious (NIID+MAL) settings. 

100 clients with varying proportions participated in the FL system, as summarized in Table \ref{tab:setting}. The malicious data rate was set at 0.2. As previously stated, each model is composed of two convolutions and two fully connected layers, with a learning rate of 0.001, an Adam optimizer, and cross-entropy as loss function. ABC-FL ran for 30 epochs, and the local models were aggregated at each epoch.

\begin{table}
    \caption{A different combination of participating clients. The first column indicates the three types of client composition, where only IID and malicious and only non-IID and malicious clients participate, respectively. Note that NIID and MAL denote non-IID and malicious, respectively. Each value is the number of clients out of a total of 100 clients. Note that the NIID rates are experimented on various proportions from 0.2 to 1.0, gradually increased by 0.2}
    \label{tab:setting}
    \centering
    \begin{tabular}{>{\centering\arraybackslash}p{2cm}|>{\centering\arraybackslash}p{1.2cm}|>{\centering\arraybackslash}p{1.2cm}|>{\centering\arraybackslash}p{1.2cm}} \hline\cline{1-4}\cline{1-4}
        \multirow{2}{*}{Setting} & \multicolumn{2}{c|}{Benign} & \multirow{2}{4em}{Malicious}\\ \cline{2-3}
                                        & IID & NIID &       \\ \hline\cline{1-4}
        \multirow{4}{*}{IID+MAL}        & 90  & 0    & 10    \\ 
                                        & 80  & 0    & 20    \\ 
                                        & 70  & 0    & 30    \\ 
                                        & 60  & 0    & 40    \\ \hline
        \multirow{4}{*}{NIID+MAL}       & 0   & 90   & 10    \\ 
                                        & 0   & 80   & 20    \\ 
                                        & 0   & 70   & 30    \\ 
                                        & 0   & 60   & 40    \\ \hline\cline{1-4}\cline{1-4}
    \end{tabular}
\end{table}

\subsection{Data \& Client Setting }

We mainly used MNIST \cite{deng2012mnist}, the most commonly used benchmark dataset in previous FL-related works \cite{shejwalkar2021manipulating, zhao2018federated, wang2019federated, geyer2017differentially}. We added the \textit{Fingers Digits 0-5} dataset \footnote{https://www.kaggle.com/roshea6/finger-digits-05} to create a non-IID dataset. The dataset consists of hand images showing a number of fingers between 0 and 5 held up and their corresponding labels from 0 to 5. Thus the data sample is of different features but bear the same label. However, since the label in the \textit{Fingers} dataset is limited to the numbers in the range of 0 to 5, we used only digits from 0 to 5 in MNIST, correspondingly.

We first randomly split the MNIST data for IID, non-IID, and malicious for each class to prepare the dataset. We prepared the IID dataset as intact digit images (MNIST) and non-IIDs as the different proportions (from 0.2 to 1 by gradually increasing the rate by 0.2) of intact finger images (Fingers) and the rest of MNIST data samples. When a non-IID rate is 0.2, for instance, the dataset for a non-IID client is prepared as a combination of fingers data samples and MNIST data samples with a 2 to 8 ratio, respectively. The malicious dataset is created by inserting the grey-cross sign (which is called a backdoor trigger) in a random location in images of ones in the MNIST dataset \cite{wu2020mitigating}. Then the target label is set to '3' for the backdoor-trigger-embedded images. We set the malicious data rate for each client as 0.2 since it has been used as the upper bound of malicious data rate in various previous studies \cite{chen2020zero, tolpegin2020data, nuding2020poisoning, cao2020fltrust}. Thus the malicious dataset consists of 20 percent of malicious data and the rest of random MNIST images.

The normal clients thereby have IID and/or non-IID images, tasked to perform image classification training only, while the malicious clients have IID and malicious images. In contrast to benign clients, the malicious clients train their models for backdoor attacks along with the main image classification task. We presented the sample images for IID, non-IID, and malicious data, as shown in Figure \ref{fig:sample-data}. 

\noindent
\textbf{Software configurations. }We used PyTorch v1.9.0 with Torchvision v0.10.0 on Python v3.8.8 and Numpy v1.20.2 for the implementation. We set seed to 42 for PyTorch, PyTorch Backends, PyTorch CUDA, and Numpy.

\noindent
\textbf{Evaluation metrics. }
We used three evaluation metrics to demonstrate our experiment results: accuracy, precision, and recall of anomalous client detection, false positive rates, and backdoor success rate before and after the detection module engages. Our proposed method does not exclude any malicious clients in the middle of the training session. Instead, the classified-as-malicious clients' model weights are not aggregated in the server while still receiving the updated global model. 

\subsection{Results}

\begin{figure}
    \centering
    \includegraphics[width=\linewidth]{"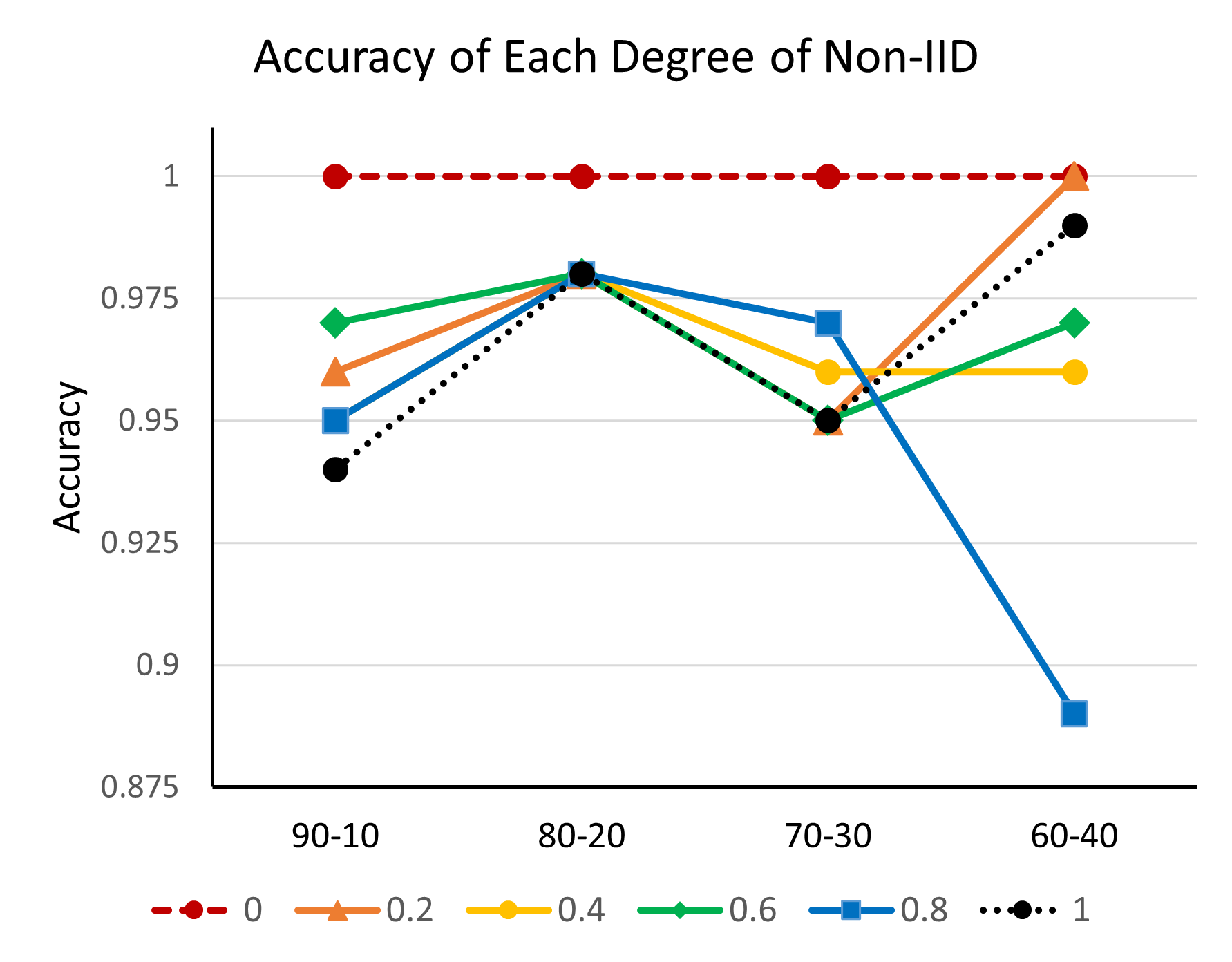"}
    \caption{Accuracy of Malicious Client Detection. The red \ding{108} dashed line, orange \ding{115} solid line, yellow \ding{108} solid line, green \ding{117} solid line, blue \ding{110} solid line, and black \ding{108} dotted line indicates the non-IID rates, 0, 0.2, 0.4, 0.6, 0.8, and 1.0 respectively. Note that the case where the non-IID rate is 0 is equivalent to the case in an IID+MAL setting. The x-axis represents the various cases, and the y-axis represents the accuracy of each case. }
    \label{fig:mal-detection-acc}
\end{figure}
\begin{figure}
    \centering
    \includegraphics[width=\linewidth]{"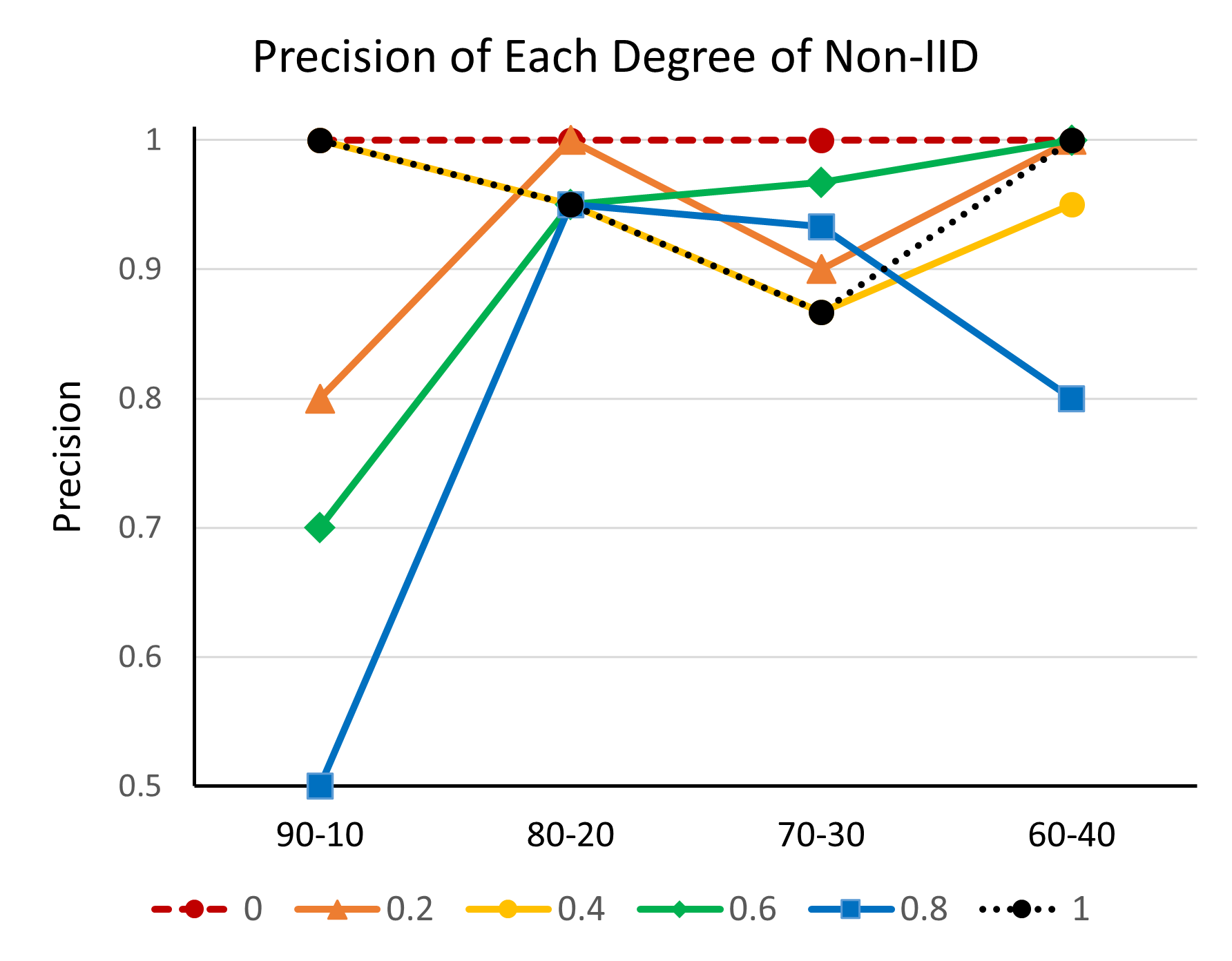"}
    \caption{Precision of Malicious Client Detection. The red \ding{108} dashed line, orange \ding{115} solid line, yellow \ding{108} solid line, green \ding{117} solid line, blue \ding{110} solid line, and black \ding{108} dotted line indicates the non-IID rates, 0, 0.2, 0.4, 0.6, 0.8, and 1.0 respectively. Note that the case where the non-IID rate is 0 is equivalent to the case in an IID+MAL setting. The x-axis represents the various cases, and the y-axis represents the precision of each case. }
    \label{fig:mal-detection-prec}
\end{figure}
\begin{figure}
    \centering
    \includegraphics[width=\linewidth]{"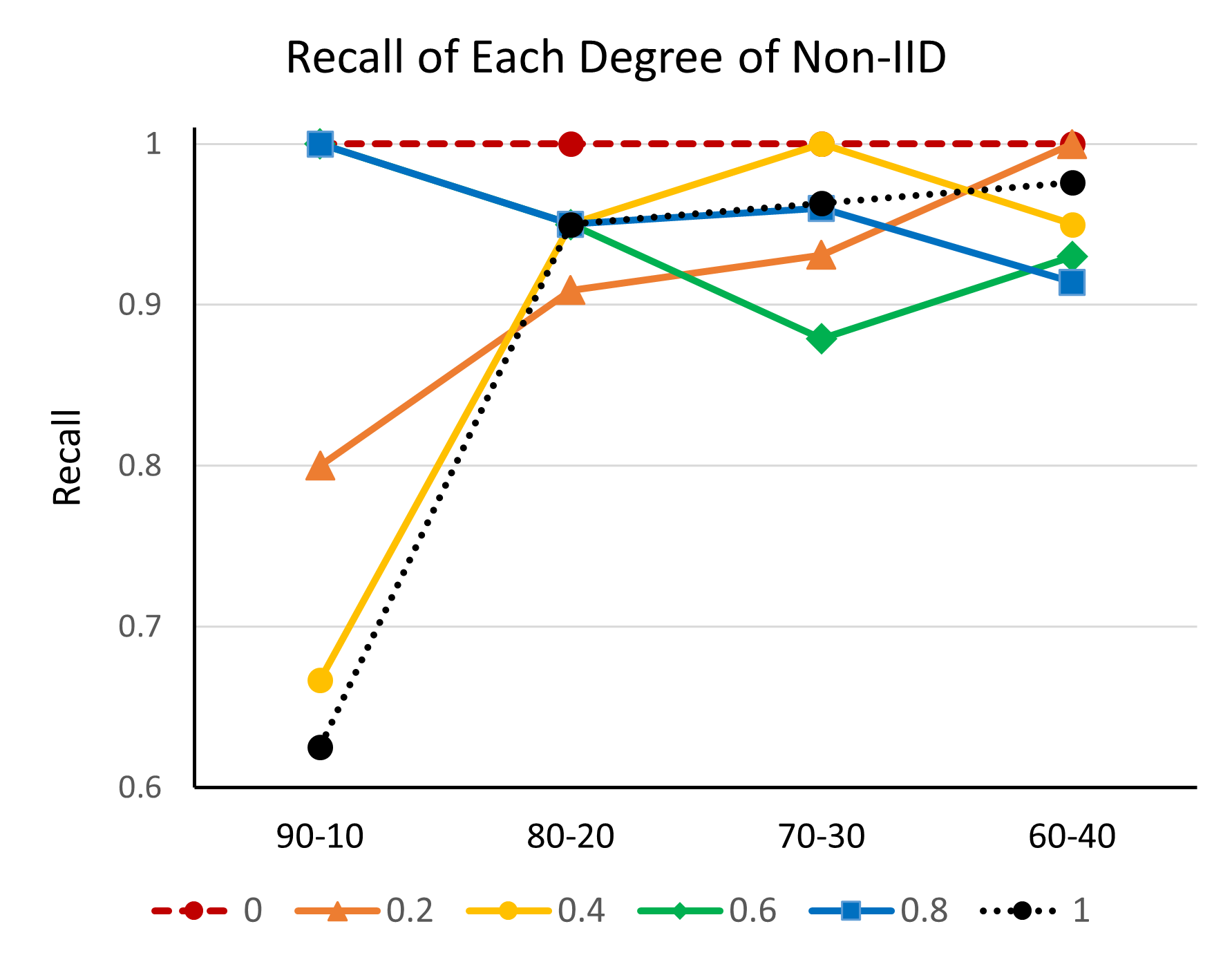"}
    \caption{Recall of Malicious Client Detection. The red \ding{108} dashed line, orange \ding{115} solid line, yellow \ding{108} solid line, green \ding{117} solid line, blue \ding{110} solid line, and black \ding{108} dotted line indicates the non-IID rates, 0, 0.2, 0.4, 0.6, 0.8, and 1.0 respectively. Note that the case where the non-IID rate is 0 is equivalent to the case in an IID+MAL setting. The x-axis represents the various cases, and the y-axis represents the recall of each case. }
    \label{fig:mal-detection-recall}
\end{figure}

We mainly experimented on three evaluation metrics: the accuracy, precision, and recall of anomalous client detection and false positive rates for each IID+MAL and NIID+MAL cases, and the decrease rate in backdoor success rate before and after applying our ABC module. In Figures \ref{fig:mal-detection-acc}, \ref{fig:mal-detection-prec}, and \ref{fig:mal-detection-recall}, the notation, such as 90-10 and 80-10, denotes the number of participating clients for each type. For example, 90-10 means that 90 NIID clients and 10 MAL clients participate, while 80-20 means that 80 NIID clients and 20 MAL clients are involved. Note that the case where non-IID degree is zero is equivalent to the cases in an IID+MAL setting.

\begin{table*}
    \caption{False Positive Rate of Malicious Client Detection for each NIID+MAL case. The setting column indicates the client combination, the case column represents the clients' proportional composition, and the non-IID rate row represents the different ratio of non-IID data to IID data that non-IID clients have, from 0.0 to 1.0, gradually increased by 0.2.} 
    \label{tab:niid_fpr}
    \centering
    \begin{tabular}{>{\centering\arraybackslash}p{1.7cm}|>{\centering\arraybackslash}p{1.3cm}|>{\centering\arraybackslash}p{1.2cm}|>{\centering\arraybackslash}p{1.2cm}|>{\centering\arraybackslash}p{1.2cm}|>{\centering\arraybackslash}p{1.2cm}|>{\centering\arraybackslash}p{1.2cm}|>{\centering\arraybackslash}p{1.2cm}} \hline\cline{1-8}\cline{1-8}
        Setting                         & Case      & \multicolumn{5}{c}{ABC Module}   \\ \hline
                                        
                      \multicolumn{2}{c|}{non-IID rate} & 0.0    & 0.2     & 0.4     & 0.6    & 0.8    & 1.0      \\ \hline\cline{1-7}
        \multirow{4}{*}{\parbox{1.2cm}{\centering NIID+MAL}}
                                        & 00-90-10      & 0.00\% & 2.22\%  & 5.56\%  & 0.00\% & 0.00\% & 6.67\%  \\
                                        & 00-80-20      & 0.00\% & 2.50\%  & 1.25\%  & 1.25\% & 1.25\% & 1.25\% \\
                                        & 00-70-30      & 0.00\% & 2.86\%  & 0.00\%  & 5.71\% & 1.43\% & 1.43\% \\
                                        & 00-60-40      & 0.00\% & 0.00\%  & 33.33\% & 3.33\% & 5.00\% & 1.67\%  \\ \hline\cline{1-8}\cline{1-8}
    \end{tabular}
\end{table*}


\subsubsection{Malicious Client Detection: Accuracy, Precision, and Recall} 
The accuracy, precision, and recall of the case where the non-IID degree is set to zero is equivalent to that of those in an IID+MAL setting. We also experimented on various non-IID compositions, assigned to each of the non-IID clients. As shown in Figure \ref{fig:mal-detection-acc}, our method is generally effective on the detection of malicious clients. Specifically, in most cases the detection accuracy approaches approximately 95\%, except for 60-40 case where the non-IID rate is 0.8, and the accuracy is 89\%. Similar trend is shown in Figure \ref{fig:mal-detection-prec} and Figure \ref{fig:mal-detection-recall}, which visualize the precision and recall, respectively. The average result of all possible cases are summarized in Table \ref{tab:avg_results}.

\subsubsection{Malicious Client Detection: FPR} 
The False Positive Rate (FPR) results were as presented in Table \ref{tab:niid_fpr}. Similarly, the FPR of the non-IID degree of zero is equivalent to that in IID+MAL setting. We also experimented on various non-IID compositions. It is notable that for most of the cases and compositions, our method does not falsely identify malicious clients. Specifically, except for the 00-60-40 where the non-IID rate is 0.4, the FPR is generally less than 7\%. Even in some cases, the FPR is 0.

\begin{table}
    \caption{The Average of the Accuracy, Precision, Recall, and FPR for all cases. } 
    \label{tab:avg_results}
    \centering
    \begin{tabular}{>{\centering\arraybackslash}p{1.5cm}|>{\centering\arraybackslash}p{1.5cm}|>{\centering\arraybackslash}p{1.5cm}|>{\centering\arraybackslash}p{1.5cm}} \hline\cline{1-4}\cline{1-4}
                                        
                    Accuracy  & Precision & Recall    & FPR       \\ \hline\cline{1-4}
                    96.94\%   & 92.71\%   & 94.80\%   & 3.56\%    \\ \hline\cline{1-4}\cline{1-4}
    \end{tabular}
\end{table}

\subsubsection{Backdoor Success Rate} 
The backdoor success rate is obtained by testing the global model on the pre-defined backdoor test dataset, consisting trigger (a gray-scaled '+' in random location) embedded data samples. Note that we prepare our dataset by inserting the trigger on MNIST digit images labeled as '1', and change the label to '3', so that the backdoor task is to learn that the trigger sign indicates the label '3'. The main task remains, classifying the number representing images. 

To evaluate the backdoor success rate, we first measure the baseline success rate. Specifically, the baseline is obtained by deploying FedAvg without the ABC module for 30 epochs for all cases described in Table \ref{tab:setting}. The backdoor success rate drops in the {IID+MAL} (as Table \ref{tab:iid_backdoor_success}) and {NIID+MAL} setting (Table \ref{tab:niid_backdoor_success}) as the ABC module is activated. Both the NIID+MAL and IID+MAL settings show similar rates of decrease. These results indicate that the ABC module not only detects participating malicious clients but significantly reduces the backdoor success rate. Table \ref{tab:iid_backdoor_success} and Table \ref{tab:niid_backdoor_success} summarize the decrease rate between the baseline and the ABC-FL for \textit{IID+MAL} and the \textit{NIID+MAL} setting, respectively.

\begin{table}
    \caption{Decrease Rate of Backdoor Success Rate before and after applying the ABC module for IID+MAL cases. Each value indicates the decrease rate from the baseline to our approach (ABC module). The setting column indicates the client combination, and the case column represents the clients' proportional composition. } 
    \label{tab:iid_backdoor_success}
    \centering
    \begin{tabular}{>{\centering\arraybackslash}p{1.7cm}|>{\centering\arraybackslash}p{1.2cm}|>{\centering\arraybackslash}p{1.7cm}} \hline\cline{1-3}\cline{1-3}
        Setting                         & Case      & {ABC Module}   \\ \hline\cline{1-3}
        \multirow{4}{*}{\parbox{1.2cm}{\centering IID+MAL}}
                                        & 90-00-10  & {84.62\%}   \\
                                        & 80-00-20  & {92.92\%}   \\
                                        & 70-00-30  & {94.27\%}   \\
                                        & 60-00-40  & {96.76\%}   \\ \hline
    \end{tabular}
\end{table}

\begin{table*}
    \caption{Decrease Rate of Backdoor Success Rate before and after applying the ABC module for NIID+MAL cases. Each value indicates the decrease rate from the baseline to our approach (ABC module). The setting column indicates the client combination, the case column represents the clients' proportional composition, and the non-IID rate row represents the different ratio of non-IID data to IID data that non-IID clients have, from 0.2 to 1.0, gradually increased by 0.2.} 
    \label{tab:niid_backdoor_success}
    \centering
    \begin{tabular}{>{\centering\arraybackslash}p{1.7cm}|>{\centering\arraybackslash}p{1.3cm}|>{\centering\arraybackslash}p{1.2cm}|>{\centering\arraybackslash}p{1.2cm}|>{\centering\arraybackslash}p{1.2cm}|>{\centering\arraybackslash}p{1.2cm}|>{\centering\arraybackslash}p{1.2cm}} \hline\cline{1-7}\cline{1-7}
        Setting                         & Case      & \multicolumn{5}{c}{ABC Module}   \\ \hline
                                        
                      \multicolumn{2}{c|}{non-IID rate} & 0.2      & 0.4      & 0.6       & 0.8      & 1.0      \\ \hline\cline{1-7}
        \multirow{4}{*}{\parbox{1.2cm}{\centering NIID+MAL}}
                                        & 00-90-10  & 85.71\%  & 100.00\% & -142.87\% & 100.00\% & 85.71\%  \\
                                        & 00-80-20  & 100.00\% & 92.86\%  & 91.43\%   & 100.00\% & 100.00\% \\
                                        & 00-70-30  & 95.48\%  & 100.00\% & 75.48\%   & 98.84\%  & 100.00\% \\
                                        & 00-60-40  & 82.18\%  & 93.99\%  & 95.99\%   & 64.37\%  & 97.77\%  \\ \hline\cline{1-7}\cline{1-7}
    \end{tabular}
\end{table*}

\section{Discussion}
\label{section5}

We experimented with various degrees of non-IIDness, from 0.0 to 1.0, gradually increasing by 0.2 because it is impossible to specify the degree of non-IIDness of the data since we cannot predict how the data differs from the IID data \cite{he2021towards}. We set an increase rate of 0.2 so as to cover as many potential real-world scenarios as possible.

We illustrate the accuracy, precision, recall of our experiment in the three figures, Figure \ref{fig:mal-detection-acc}, Figure \ref{fig:mal-detection-prec}, and Figure \ref{fig:mal-detection-recall}, respectively. Note that the result for the IID+MAL setting is equivalent to the case where the non-IID rate is zero. 

Accuracy is the ratio of the total number of correct predictions to the total number of predictions, as shown in Eq. \ref{eq:acc}.
\begin{equation}
accuracy = \frac{TP + TN}{TP + FN + TN + FP}
\label{eq:acc}
\end{equation}

Precision is the ratio of the true positives to all the positives, as shown in Eq. \ref{eq:precision} In other words, the precision is a measure of the malicious clients that we correctly identify as anomalies out of all the clients that pose to be malicious.
\begin{equation}
precision = \frac{TP}{TP + FP}
\label{eq:precision}
\end{equation}

Recall is a ratio of the true positives to the sum of the true positives and the false negatives, as shown in Eq. \ref{eq:recall};
The recall indicates how many clients we correctly identified as anomalies for all the actual malicious clients. 
\begin{equation}
recall = \frac{TP}{TP + FN}
\label{eq:recall}
\end{equation}

Figures \ref{fig:mal-detection-acc}, \ref{fig:mal-detection-prec}, and \ref{fig:mal-detection-recall} illustrate our experimental results which indicate that our proposed method identifies malicious clients with high confidence in the NIID+MAL setting with a varying degree of non-IIDness. The only exception would be, for the case of 00-90-10, where the 90 NIID clients and 10 MAL clients participate in the FL procedure, the precision and recall are relatively lower than the other cases. This is due to the relatively small portion of malicious clients' present. Nevertheless, the high accuracy, slightly compensates for the defects. 

The Table \ref{tab:niid_fpr} shows the False Positive Rates for various NIID+MAL cases, with varying non-IID rates. 
\begin{equation}
FPR = \frac{FP}{FP + FN} \times 100 (\%)
\end{equation}
As mentioned above, the FPR for the IID+MAL setting is equivalent to the case where the non-IID rate is 0. Observably, even for setting that the non-IID rate is not 0, the FPR is low (average FPR is 3.56\% as Table \ref{tab:avg_results}). We assume this is due to the positive influence of the clustering method. Through dynamic hierarchical clustering, we determined the maliciousness of the clients on the unit of clusters instead of treating them as individual points. The FPR, however, is 33.33\% in the 00-60-40 case when the non-IID rate is 0.4. This is a side effect of the clustering basis, as clusters are falsely identified as malicious so that the clients belonging to those clusters are also incorrectly classified as malicious. Nevertheless, in general, the clustering-based identification shows a low rate of FPR, proving that our method is effective in identifying malicious clients.

As summarized in the Table \ref{tab:avg_results}, the average accuracy is 96.94\%, precision as 92.71\%, recall as 94.80\%, and the FPR is 3.56\%. The aforementioned average is the average of all cases in IID+MAL and NIID+MAL settings. Based on these results, we can calculate the F1 score, which is the harmonic mean of the precision and recall. The F1 score of our average result is 93.74. This indicates that our proposed method is generally well-balanced between the precision and recall trade-off and robustly and effectively detects malicious clients, even though it shows relatively lower precision and recall in few isolated cases.

In Table \ref{tab:iid_backdoor_success}, we examined the backdoor success rate before and after applying the ABC module in various IID+MAL settings. The backdoor accuracy decreased on average by 92.14\%. 
Table \ref{tab:niid_backdoor_success} summarizes the backdoor success rate before and after applying the ABC module in various NIID+MAL settings. The detection rates declined by a significant amount, except for the 00-90-10 case where the non-IID rate was 0.6. For that certain case, the attack success rate increased instead of decreasing. We assume that this is due to the low backdoor success rate in the baseline. As we stealthily embedded the backdoor trigger, the backdoor attack was not successful in such low proportional malicious client participation. Nevertheless, the reduction in backdoor success rate is on average 80.85\%; the results validate that our proposed method detects the malicious clients and decreases the negative impact of the malicious clients by excluding their model parameters when updating the global model. \\

\noindent\textbf{Future works.} In future works, we will experiment our proposed method on malicious client rates exceeding 0.5 as where in this work, we assumed that fewer than half of the malicious clients were active. This is because we treat the system as compromised if the malicious client occupies the majority, in which case it should be abandoned and re-initialized. Aside from our premise, it would be more beneficial to detect malicious clients even if they constitute a majority of the population. 
Furthermore, we will examine our method on multiple backdoor tasks. The proposed method in this study is only tested on a single backdoor task, where we inserted a backdoor trigger and flipped the label from 1 to 3. The attackers, however, may try to inject multiple backdoor tasks with differently embedded backdoor triggers. Thus, we will examine our method on various backdoor tasks in the future. 
We also leave the validation of our proposed method through various malicious data rates to our upcoming research. In this study, we set the malicious data rate as 0.2 with varying malicious client rates. Given that the malicious client's objective is to avoid detection, we determined that a lesser malicious data rate, which can be more stealthy, would be a more reasonable assumption. However, as we cannot assume the degree of attacks, we will experiment with varying malicious data rates in the future.

\section{Conclusion}
\label{section6}

FL, a distributed machine learning framework, is increasingly gathering attention as it neither collects nor has access to clients' raw data. Due to the restricted access to the raw data, however, it is challenging to classify malicious clients when the clients with non-IID data distributions participate; ; nevertheless, it is more natural to assume that real-world data are generally coming in a non-IID fashion. Our proposed ABC module mitigates anomalous clients whose objective is to backdoor our system even when clients with non-IID data exist. We use feature dimension reduction, dynamic clustering, and cosine similarity-based clipping to identify and classify these three types of clients in this work. Our experiment results demonstrate a successful anomalous client detection with an accuracy of 96.94\% and a false positive rate of 3.56\%, on average. We also demonstrate that our model significantly reduces the influence of anomalous clients. Our findings may be helpful in future research that aim to effectively eliminate fraudulent clients while simultaneously training a model with a varied set of data.

\section*{Acknowledgements}
We thank the anonymous reviewers for their insightful reviews.

\bibliographystyle{ACM-Reference-Format}
\bibliography{Reference}

\end{document}